\documentclass[sn-mathphys,Numbered]{sn-jnl}

\usepackage{graphicx}%
\usepackage{multirow}%
\usepackage{amsmath,amssymb,amsfonts}%
\usepackage{amsthm}%
\usepackage{mathrsfs}%
\usepackage[title]{appendix}%
\usepackage{xcolor}%
\usepackage{physics}%
\usepackage{textcomp}%
\usepackage{manyfoot}%
\usepackage{booktabs}%
\usepackage{algorithm}%
\usepackage{algorithmicx}%
\usepackage{algpseudocode}%
\usepackage{listings}%
\usepackage{siunitx}%
\usepackage{verbatim}
\usepackage{subcaption}

\theoremstyle{thmstyleone}%
\theoremstyle{thmstyletwo}%

\theoremstyle{thmstylethree}%

\begin{document}

\title{Mechanics and Design of Metastructured Auxetic Patches with Bio-inspired Materials}


\author*[1]{\fnm{Yingbin} \sur{Chen}}\email{yingbin-chen@uiowa.edu}

\author[2]{\fnm{Milad} \sur{Arzani}}\email{milad-arzani@uiowa.edu}
\equalcont{These authors contributed equally to this work.}

\author[2]{\fnm{Xuan} \sur{Mu}}\email{xuan-mu@uiowa.edu}
\equalcont{These authors contributed equally to this work.}

\author[3]{\fnm{Sophia} \sur{Jin}}\email{sophia-jin@uiowa.edu}
\equalcont{These authors contributed equally to this work.}

\author[1]{\fnm{Shaoping} \sur{Xiao}}\email{shaoping-xiao@uiowa.edu}
\equalcont{These authors contributed equally to this work.}

\affil[1]{\orgdiv{Department of Mechanical Engineering, Iowa Technology Institute}, \orgname{University of Iowa}, \orgaddress{\street{3131 Seamans Center}, \city{Iowa City}, \postcode{52242}, \state{Iowa}, \country{USA}}}

\affil[2]{\orgdiv{Roy J. Carver Department of Biomedical Engineering}, \orgname{University of Iowa}, \orgaddress{\street{5601 Seamans Center}, \city{Iowa City}, \postcode{52242}, \state{Iowa}, \country{USA}}}

\affil[3]{\orgname{Liberty High School}, \orgaddress{\street{1400 S Dubuque St}, \city{North Liberty}, \postcode{52217}, \state{Iowa}, \country{USA}}}


\abstract{Metastructured auxetic patches, characterized by negative Poisson's ratios, offer unique mechanical properties that closely resemble the behavior of human tissues and organs. As a result, these patches have gained significant attention for their potential applications in organ repair and tissue regeneration. This study focuses on neural networks-based computational modeling of auxetic patches with a sinusoidal metastructure fabricated from silk fibroin, a bio-inspired material known for its biocompatibility and strength. The primary objective of this research is to introduce a novel, data-driven framework for patch design. To achieve this, we conducted experimental fabrication and mechanical testing to determine material properties and validate the corresponding finite element models. Finite element simulations were then employed to generate the necessary data, while greedy sampling, an active learning technique, was utilized to reduce the computational cost associated with data labeling. Two neural networks were trained to accurately predict Poisson's ratios and stresses for strains up to 15\%, respectively. Both models achieved $R^2$ scores exceeding 0.995, which indicates highly reliable predictions. Building on this, we developed a neural network-based design model capable of tailoring patch designs to achieve specific mechanical properties. This model demonstrated superior performance when compared to traditional optimization methods, such as genetic algorithms, by providing more efficient and precise design solutions. The proposed framework represents a significant advancement in the design of bio-inspired metastructures for medical applications, paving the way for future innovations in tissue engineering and regenerative medicine.}

\keywords{metastructure; auxetic patches; silk fibroin; neural networks}



\maketitle

\section{Introduction}\label{sec1}

Unlike traditional engineering materials, human organs such as the lung  \cite{Tomoda2013}, heart \cite{Ferraiuoli2019}, and stomach \cite{Banerjee2020} exhibit unique auxetic deformation behavior, characterized by negative Poisson's ratios. In other words, while conventional materials contract transversely when stretched longitudinally, these organs expand in directions perpendicular to the applied force. This mechanical property is vital for the dynamic and complex functions of biological tissues, helping them withstand and recover from a wide range of mechanical stresses. For medical applications, particularly in organ repair and regeneration, therapeutic patches are expected to replicate this auxetic behavior to mimic the natural properties of these tissues \cite{Mir2014}. This property is essential for ensuring seamless integration with surrounding tissue and proper function under physiological conditions. Consequently, developing bioengineered patches with auxetic characteristics presents a promising strategy to enhance the success of organ regeneration therapies. 

Significant progress has been made in designing, fabricating, and testing therapeutic patches, with extensive efforts focused on incorporating auxetic features into synthetic materials. Among these structures, the re-entrant honeycomb design has been widely studied as a type of metastructured auxetic material \cite{Li2018, Kapnisi2018}. Other metastructures include chiral truss, orthogonal oval voids, pinwheel designs, arrowhead, and others \cite{Chansoria2022a}. Olvera \textit{et al.} \cite{Olvera2020} fabricated an electroconductive cardiac path using melt electrospinning writing (MEW). Unlike conventional square patch designs, these auxetic patches, featuring missing rib unit cells, accommodated the deformation patterns of the human myocardium. The researchers tested the patches' Poisson's ratio in transverse and longitudinal directions, observing a negative Poisson's ratio up to 40\% strain in the transverse direction and up to 60\% in the longitudinal direction. In another work, Brazhkina \textit{et al.} \cite{Brazhkina2021} created auxetic cardiac patches from polycaprolactone and gelatin methacrylate with various thicknesses through 3D printing. These patches supported the function of cardiomyocytes derived from induced pluripotent stem cells over a 14-day culture period, demonstrating their therapeutic potential for cardiovascular diseases. Similarly, Asulin \textit{et al.} \cite{Asulin2021} utilized 3D printing to fabricate dog-bone-shaped bi-layer patches integrated with soft and stretchable electronics for cardiac applications. For a broader discussion of similar research, see the following review papers \cite{Lecina2023, Rose2023, Mei2020}.

Bio-inspired materials have recently gained considerable attention for their biocompatibility, sustainability, and environmentally friendly characteristics, positioning them as promising candidates for auxetic patch fabrication. Unlike plastics and metals, structural proteinaceous materials - such as collagen, fibrinogen, and silk fibroin - have been increasingly adopted in the development of metastructured auxetic materials \cite{Chansoria2024, Mardling2020, Pradhan2020, Shu2021}, owing to their inherent biochemical properties and mechanical performance \cite{Abascal2018, Harrington2021, Hu2012, Miserez2023, Sorushanova2019}. Among these materials, silk fibroin stands out for its customizable mechanical properties and prolonged degradation profile, which can last from months to years. Silk fibroin, derived from domestic \textit{Bombyx mori} silkworms \cite{Mu2020a, Rockwood2011}, has been utilized in a wide range of biomedical applications, including wound dressings \cite{Zhang2020b} and implantable devices \cite{Guo2020b, Li2020}. Notably, it has led to the development of several FDA-approved clinical products \cite{Holland2019}. 

The mechanical properties, biocompatibility, and degradability of silk fibroin-based metamaterials are primarily influenced by the fabrication process and the arrangement of polypeptide chains, including secondary structures and nanofibrils \cite{Guo2020a, Mu2022b, Koh2015}. Various factors, such as temperature change \cite{Guo2020b}, sonication \cite{Guziewicz2011}, and organic solvents \cite{Zhu2018}, can influence the assembly of silk fibroin molecules during fabrication. However, these factors can compromise the structural integrity, biocompatibility, and biodegradability of silk fibroin and other proteins \cite{Gupta2022}, limiting the full potential of silk fibroin-based auxetic metamaterials. One promising approach involves using mild conditions, such as aqueous salt solutions, to process silk fibroin into structural materials in 3D printing \cite{Mu2022a, Mu2020b} and casting \cite{Kim2005}. While the detailed mechanism is still under investigation, this approach may mimic the native spinning conditions observed in silkworms and web-weaving spiders \cite{Andersson2016, Mu2023}. This strategy holds significant promise for advancing the development and fabrication of metastructured auxetic silk fibroin materials. 

Beyond the fabrication and testing of metamaterials, advancing this field requires tackling two critical and interconnected challenges in the design process. The first challenge lies in accurately predicting the mechanical properties of designed auxetic patches, including key parameters such as Poisson's ratios and stresses across a range of deformation. This is critical due to the inherently nonlinear behavior of auxetic materials, which renders sign-point material property prediction less effective. Reliable predictive models are essential to ensure that these materials perform as intended under real-world conditions, especially in applications involving complex and dynamic environments. The second challenge involves designing patches with highly specific and targeted mechanical behaviors tailored to meet the unique functional demands of various organs or application contexts. Achieving this requires a nuanced understanding of both the material properties and the biomechanical requirements of the target system, as well as the ability to translate these needs into precise material configurations. These challenges are deeply intertwined during the design process. Accurate predictive modeling informs targeted designs, while the complexity of designing for specificity underscores the critical importance of reliable predictions. Successfully addressing these challenges would significantly reduce the reliance on the current trial-and-error approach in experiments, streamline the iterative design process, and ultimately accelerate progress in the field. 

Finite element methods (FEMs) \cite{Ted2014FEM} have been widely used to model and simulate the mechanical properties of auxetic patches and to assist in their design. Chansoria \textit{et al.} \cite{Chansoria2020a, Chansoria2019a, Chansoria2019b} employed multiphysics FEM modeling to support the experimental design and characterize the effects of process parameters. Their study screened over 116 patch designs, culminating in an Ashby chart that illustrated the correlation between Poisson's ratio and structure stiffness \cite{Chansoria2022b}. Similarly, Javadi \textit{et al.} \cite{Javadi2012} integrated FEM with a genetic algorithm (GA) to design and optimize the microstructures of auxetic materials. In another notable work, Meier \textit{et al.} \cite{Meier2024} introduced a systematic design approach that combined FEM, GA, and automated modeling to create metamaterial lattice structures with desired isotropic and auxetic properties. The optimal structures were fabricated and tested to validate their performance experimentally. Zhang \textit{et al.} \cite{Zhang2024} also used FEM to verify the optimal anti-chiral parallelogram configurations derived from GA optimizations and theoretical models.  It is worth noting that most of these studies assumed that the metastructured patches and their constituent materials exhibited linear elastic behavior.

In the era of artificial intelligence (AI), advanced machine learning (ML) technologies, such as deep learning (DL), have been increasingly applied to assist in modeling and simulation \cite{Xiao2023a, Faruk2024}. Artificial neural networks (ANNs), in particular, have proven to be valuable tools in the design of metamaterials. Yan and co-workers \cite{Yan2024} trained neural networks to provide computationally inexpensive predictions of the auxeticity and stiffness of porous metamaterials. These networks generated a comprehensive database that enabled parametric analyses to explore the relationship between material auxeticity and stiffness and supported inverse design processes to optimize metastructures for these properties. Similarly, Shen \textit{et al.} \cite{Shen2024} employed ANNs as surrogates for FEM to predict mechanical properties. They further utilized a generative adversarial network (GAN) in an inverse optimization framework based on datasets generated by the ANN regression model. In another study, Mohammadnejad \textit{et al.} \cite{Mohammadnejad2024} used FEM to simulate metamaterials with 1500 different sets of structure parameters. The resulting dataset was then utilized to train ANNs, which served as a robust tool for precisely designing metamaterials to achieve desired stress-strain and Poisson's ratio-strain behaviors. 

Notably, these studies primarily utilized material properties as inputs and design parameters as outputs for training ANNs in the inverse design process. However, Peng and Xue \cite{Peng2024} highlighted that the inverse design task is inherently ill-posed, as the optimal design might not be unique for a given set of desired properties. To address this limitation, they introduced an additional term in the ANN loss function to evaluate the error between the predicted and actual design variables, building on the methodologies in \cite{Kumar2020, Bastek2022}. This term was activated only during the initial phase of ANN training to aid convergence, thereby improving the model's ability to resolve the ambiguity in the inverse design process effectively.

This paper adopts a data-driven approach, leveraging ANNs as surrogate models to address the previously mentioned challenges in the mechanics and design of auxetic patches made from bio-inspired materials like silk fibroin. Experimental testing was conducted not only to characterize the stress-strain relationship of silk fibroin for numerical modeling but also to validate the FEM model of the metastructured patches. Data for this study was collected through FEM simulations. This work considers the nonlinear mechanical behavior of metastructured patches, which differs from most existing studies that assumed linear elasticity. This approach more accurately reflects the mechanics of organs, allowing for potential future applications. Consequently, ANNs were trained to predict not just single values but a series of Poisson's ratios and stresses at various strain levels. Additionally, unlike previous works \cite{Yan2024, Mohammadnejad2024}, this study employed an active learning technique \cite{Chen2024} to significantly reduce the number of FEM simulations required for data labeling. 

Another major contribution involves developing a novel framework for training ANNs as design models to optimize metastructured patches tailored to specific (i.e., desired) mechanical properties. Unlike existing studies, our approach relied solely on material properties from the collected dataset during training. The predicted design variables were fed into the ANN mechanical property prediction models to compare with the targeted properties and compute the baseline loss function. To address the inherent ill-posed nature of inverse design processes, we revised the baseline loss function differently from previous studies \cite{Peng2024}. Rather than introducing regularization terms to align predictive designs with actual ones, we eliminated designs with proportional design variables by incorporating additional loss terms. As a result, our design model can predict multiple diverse optimal designs as needed. This work provides a valuable toolset for organ repair and regeneration, streamlining the design, fabrication, and testing of therapeutic patches for future applications. 

This paper is organized as follows. After the introduction, we describe the FEM modeling and simulation of metastructured patches. The use of ANNs for predicting mechanical properties and designing metastructures is also discussed. The experimental testing section follows, detailing the fabrication and testing processes, as well as the validation of numerical simulations. Next, we present and discuss the performance of ANNs in studying the mechanics and design of metastructures while also addressing the limitations of this work. Finally, the conclusion summarizes our findings and outlines future research.

\section{Numerical modeling and machine learning}\label{sec2}

\subsection{Patch structure and finite element modeling}\label{sec21}

Auxetic materials exhibit a distinctive negative Poisson's ratio, making them well-suited for creating patches to match the unique mechanical behavior of human organs. A variety of macroscopic metastructures display auxetic properties, including re-entrant structures \cite{gibson1982}, rotating polygonal structures \cite{grima2000}, chiral structures \cite{lakes1991}, crumpled sheet structures \cite{alderson2012}, and perforated sheet structures \cite{grima2010}. Notably, the re-entrant category includes several configurations, such as hexagonal honeycomb \cite{burton1997}, star honeycomb \cite{grima2005}, triangular \cite{zhang2022}, and sinusoidal structures \cite{Dolla2007}. These metastructures demonstrate that auxetic behavior is strongly influenced by geometric parameters, highlighting the importance of geometric design in customizing auxetic patches.

In this study, we choose the re-entrant sinusoidal metastructure to design auxetic patches, as this structure can be defined by just three design variables and precisely represented through mathematical formulas. This streamlined configuration is particularly advantageous for ML applications. Labeling training data from FEM simulations is typically a time-intensive task; however, the limited number of design variables allows for a smaller training dataset, thereby reducing the time required for data generation. Furthermore, the formulaic representation of the structure enhances the construction of numerical models through parameterization, greatly simplifying the model preparation process.

\begin{figure}[ht]
    \centering
    \includegraphics[width=0.65\textwidth]{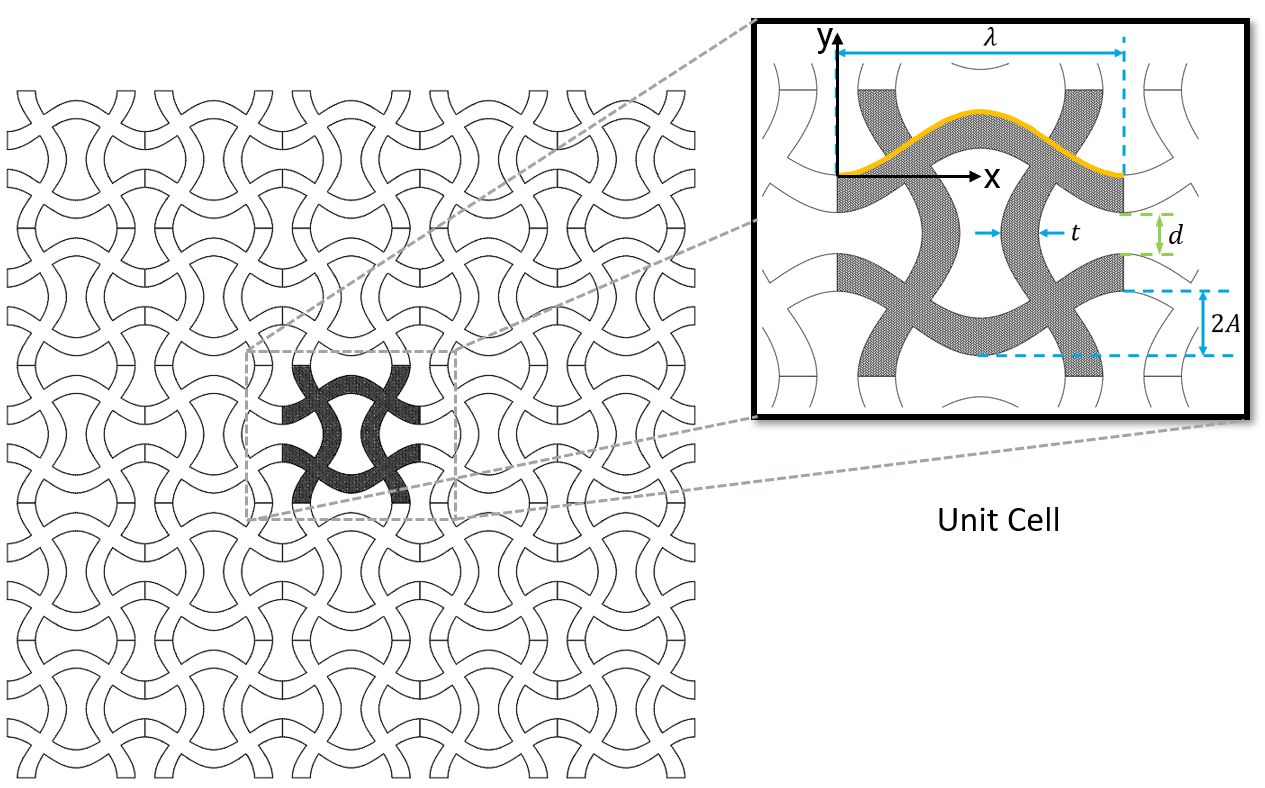}
    \caption{Sinusoidal metastructure.}
    \label{fig:Sinusoidal_structure}
\end{figure}

The re-entrant sinusoidal structure is depicted in Figure~\ref{fig:Sinusoidal_structure}. For the unit cell, we use the top sinusoidal curve (shown in yellow) as an example, with its left endpoint positioned at the coordinates $[0,0]$. This curve is mathematically defined as:
\begin{equation}\label{eq:sinusoidal_structure}
y(x) = A \sin\left(\frac{2\pi x}{\lambda}\right)
\end{equation}
where $A$ represents the amplitude and $\lambda$ is the wavelength. By applying translation and rotation operations, we can derive the remaining seven curves that compose the structure. The thickness of the ligament, denoted by $t$, is defined during the translation process. These three parameters ($A$, $\lambda$, and $t$) are the design variables of the sinusoidal metastructures in this study. Notably, the distance $d$ between the peaks is automatically determined by the structure's periodicity:
\begin{equation}\label{eq:design_function}
d = \frac{\lambda}{2} - 2A - t
\end{equation}

To further reduce the training dataset size, we employ the Greedy Sampling (GS) method, an active learning strategy previously validated in our research for enhancing ML efficiency in the materials science field \cite{Chen2024}. GS is a pool-based active learning method that requires defining an initial pool containing all potential samples. In this context, the pool of unlabelled data is established by randomly varying the three design variables: amplitude ($A$) from 0.2 mm to 2.1 mm, wavelength ($\lambda$) from 2 mm to 21 mm, and thickness of the ligament ($t$) from 0.2 mm to 2.1 mm. Then, the GS method selects up to 150 of the most informative data from this pool to conduct FEM simulations for labeling.

\begin{figure}[ht]
    \centering
    \includegraphics[width=0.5\textwidth]{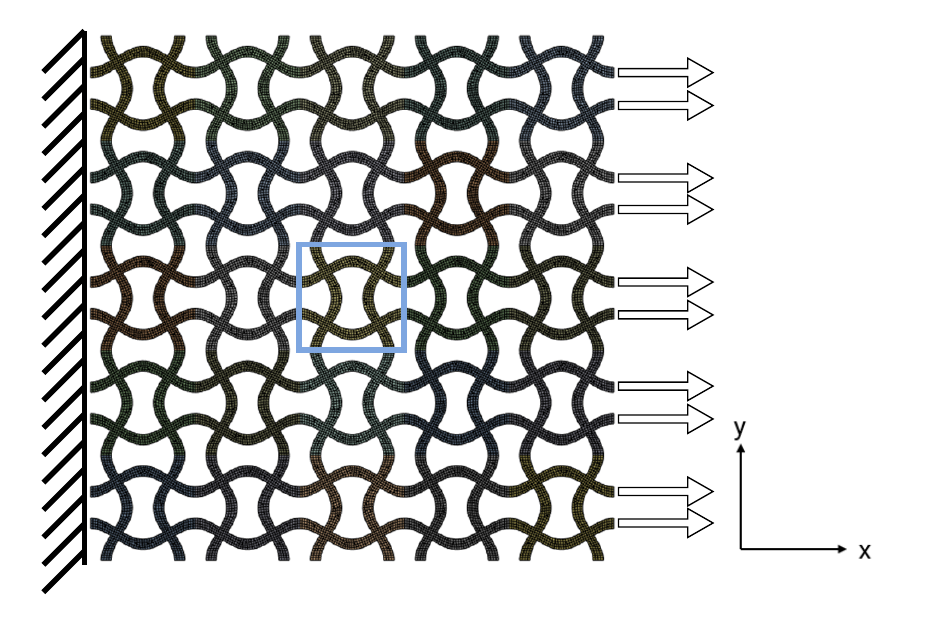}
    \caption{FEM simulation diagram.}
    \label{fig:FEM_boundary}
\end{figure}

After selecting a specific design, we replicate the unit cell into a $5 \times 5$ array to construct the complete patch structure. FEM simulations are performed using ANSYS Workbench, version 2023 R1. As shown in Figure~\ref{fig:FEM_boundary}, we fix the degrees of freedom (DOF) in the x-direction on the left side and apply an x-direction displacement on the right side to induce a nominal strain up to $15\%$ (i.e., $0.15 \times 5\lambda$). The SHELL 181 element is utilized, with a uniform element thickness $1$ mm. Material properties are obtained via experimental testing and will be discussed in Section~\ref{sec32}.

Simulations are conducted on the entire patch; however, our stress and strain analyses specifically target the central unit cell (indicated by the blue box in Figure~\ref{fig:FEM_boundary}), to obtain its mechanical properties at various strains, including Poisson's ratios $\nu$ and nominal stresses $\sigma$. The Poisson's ratio is defined as:
\begin{equation}\label{eq:poi_define}
\nu = \frac{-\varepsilon_y}{\varepsilon_x}
\end{equation}
where $\varepsilon_x$ and $\varepsilon_y$ represent the nominal strains in the $x$- and $y$-directions, respectively. The nominal stress is calculated using:
\begin{equation}\label{eq:stress_defin}
\sigma = \frac{F_R}{5\lambda t_e}
\end{equation}
where, $F_R$ is the reaction force resulting from the applied displacement. This approach enables us to adopt periodic boundary conditions, ensuring data stability and consistency for use in ML applications. 

\subsection{ANNs for regression}\label{sec22}

As a prominent branch of supervised ML, DL is increasingly applied in materials science \cite{Xiao2023a}, enhancing modeling and simulation efforts. Its ability to learn complex relationships between variables makes it particularly suitable for predicting material properties and designing new materials, especially for exploring the intricate connections between design parameters and the mechanical properties of metastructures in our research. In this study, we employ ANNs with fully connected layers to predict the mechanical properties of metastructures based on their design parameters. Building on these predictive models, we propose a novel data-driven design framework to train ANNs (serving as design models) to generate the design parameters of metastructures according to specified mechanical properties. 

\begin{figure}[ht]
    \centering
    \includegraphics[width=0.45\textwidth]{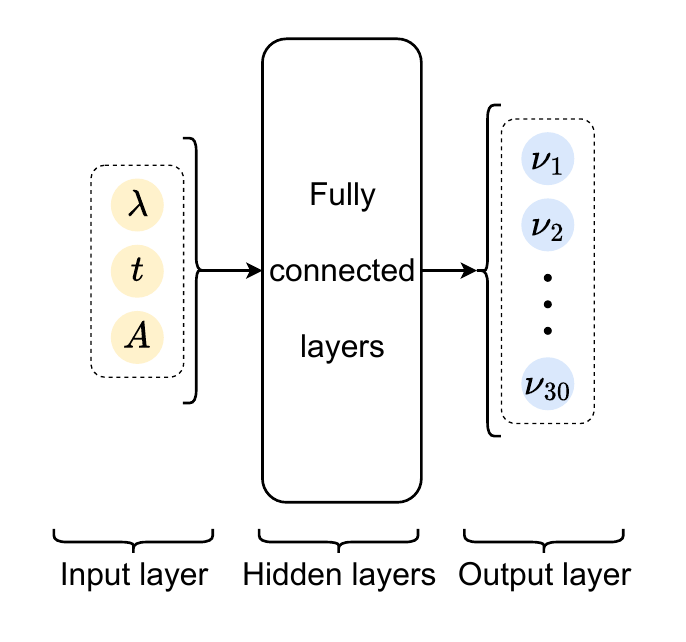}
    \caption{The architecture of an ANN to predict Poisson's ratios.}
    \label{fig:predictive}
\end{figure}

Two ANNs have been developed and trained to predict Poission's ratios and stresses, respectively, across a range of strain values from 0.5\% to 15\%, at 0.5\% intervals. Both networks share the same architecture, as illustrated in Figure~\ref{fig:predictive}. The input layer consists of three neurons corresponding to the design variables: $\lambda$, $t$, and $A$, as discussed in the previous subsection. The output layer features 30 neurons, each representing the predicted value at a specific strain. Additionally, the architecture includes four hidden fully connected layers with 50, 100, 125, and 75 neurons, respectively. The predictive models utilize the Rectified Linear Unit (ReLU) as the activation function. During training, the models minimize the mean squared error (MSE) loss function while employing the Adam optimizer. 

The data is collected from 150 FEM simulations selected based on the GS strategy. It is segmented into three sets: training (132 data samples), validation (9 data samples), and testing (9 data samples). The validation set is used to tune hyperparameters, including the number of layers and neurons, learning rate, and number of epochs. Additionally, the test set is utilized to evaluate the models' performance using the $R^2$ score metric.

\subsection{ANN for design}\label{sec23}

We aim to design metastructured patches with mechanical properties that approximate specified targets. In this study, an individual sinusoidal metastructure design is defined by the parameters $\lambda$, $t$, and $A$, while the target properties consist of Poisson's ratios and stresses at various strains. Interestingly, other designs with parameters $n\lambda$, $nt$, and $nA$ (where $n = 1, 2, 3 ... $) can exhibit same properties. Consequently, using a conventional ANN to compute the loss between the predicted design variables and the targets does not fully achieve our objectives. Inspired by Cycle-Consistent Adversarial Networks \cite{Zhu2017}, which address image-to-image translation tasks without paired training data, we propose a novel approach that incorporates a cycle-consistent-type loss within an ANN framework to recommend design(s) for desired mechanical properties. 

\begin{figure}[ht]
    \centering
    \includegraphics[width=0.85\textwidth]{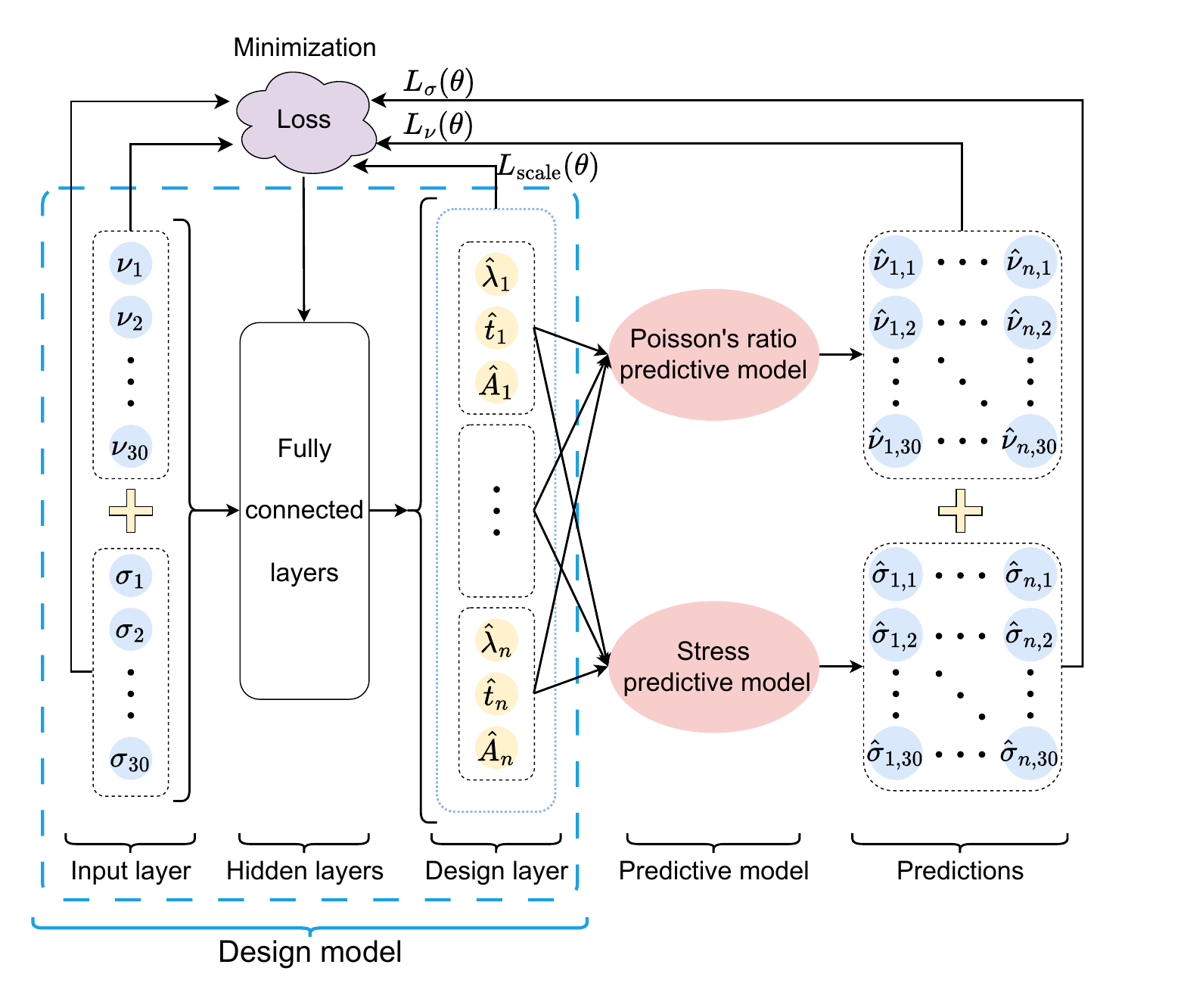}
    \caption{The schematic diagram of the design process.}
    \label{fig:Design_model}
\end{figure}

The architecture of our design model and its training process is depicted in Figure~\ref{fig:Design_model}. Like conventional ANN structures, the design model comprises an input layer, several hidden layers, and a design layer serving as the output layer. However, the dataset consists of only inputs (i.e., desired Poisson's ratios and stresses) without corresponding outputs (i.e., design parameters). Instead, the design layer is integrated with high-performance predictive models (developed in Section \ref{sec22}) that provide feedback on the mechanical properties of proposed designs generated in the design layer.  

Specifically, the targeted mechanical properties include Poisson's ratios and stresses at 30 strain levels, ranging from 0\% to 15\% in 0.5\% intervals. As a result, the input layer consists of 60 neurons, as shown in Figure~\ref{fig:Design_model}. The hidden layer module includes five fully connected layers configured with 90, 125, 150, 100, and 50 neurons, respectively, each utilizing the ReLU activation function. The design layer is flexible to contain $3n$ neurons grouped into $n$ sets of design parameters, $(\hat{\lambda}_1, \hat{t}_1, \hat{A}_1), ... (\hat{\lambda}_n, \hat{t}_n, \hat{A}_n)$, to represent different design schemes. These parameters are then passed to predictive models to forecast Poisson's ratios and stresses. The loss function is calculated by comparing the desired properties with these predictions. To promote design diversity, we incorporate an additional term into the model, enhancing the original cycle-consistent loss. Consequently, the model minimizes the total loss function and updates the weights with the Adam optimizer.

We define the loss function $L(\theta)$ in the following equation:
\begin{equation}\label{eq:loss_define}
L(\theta) = \alpha L_{\nu}(\theta) + \beta L_{\sigma}(\theta) + \gamma L_{\text{scale}}(\theta)
\end{equation}
\begin{equation}\label{eq:loss_nu}
L_{\nu}(\theta) = \frac{1}{NQ} \sum_{i=1}^N \sum_{j=1}^Q (\nu_j - \hat{\nu}_{ij}(\theta))^2
\end{equation}
\begin{equation}\label{eq:loss_sigma}
L_{\sigma}(\theta) = \frac{1}{NQ} \sum_{i=1}^N \sum_{j=1}^Q (\sigma_j - \hat{\sigma}_{ij}(\theta))^2
\end{equation}
\begin{equation}\label{eq:loss_scale}
L_{\text{scale}}(\theta) = \frac{1}{\frac{1}{P \times \binom{N}{2}} \sum_{\substack{i, j = 1 \\ i < j}}^N \sum_{k=1}^P \left(\frac{\hat{D}_{ik}\left(\theta\right)}{\hat{D}_{jk}\left(\theta\right)} -\hat{M}_{ij}\left(\theta\right)\right)^2}
\end{equation}
\begin{equation}\label{eq:loss_scale_}
\hat{D}_{i1}\left(\theta\right), \hat{D}_{i2}\left(\theta\right), \hat{D}_{i3}\left(\theta\right) = \hat{\lambda}_i\left(\theta\right), \hat{t}_i\left(\theta\right), \hat{A}_i\left(\theta\right)
\end{equation}
\begin{equation}\label{eq:loss_scale_}
\hat{M}_{ij}\left(\theta\right) = \frac{1}{P} \sum_{\substack{k = 1}}^P \frac{\hat{D}_{ik}\left(\theta\right)}{\hat{D}_{jk}\left(\theta\right)}
\end{equation}
where $\theta$ encompasses all neural network weights and biases, and $\alpha$, $\beta$, and $\gamma$ act as tunable hyperparameters that balance the contributions of each loss component. $N$ represents the number of distinct design groups, $Q$ (= 30) denotes the number of strain levels where Poisson's ratios and stresses are evaluated, and $P$ (= 3) corresponds to the number of design parameters within each sinusoidal metastructure.

The total loss function comprises three components. The terms $L_{\nu}(\theta)$ and $L_{\sigma}(\theta)$ ensure that the predicted mechanical properties ($\hat{\nu}_i, \hat{\sigma}_i$) closely align with the desired values ($\nu_i, \sigma_i$) fed to the input layer. The third component $L_{\text{scale}}(\theta)$ penalizes designs that are merely scaled versions of each other, enabling diverse design recommendations. In Eqn~(\ref{eq:loss_scale}), $\hat{D}_{ij}$, where $i=1, ..., N$ and $j=1, ..., P$, represents the recommended design variables. The term $\hat{M}_{ij}\left(\theta\right)$ represents the average ratio of design parameters between the $i$-th and $j$-th design groups.

We use the dataset generated from the same simulation results described in Section~\ref{sec21} to train, validate, and test the design model. A similar approach is employed to tune the hyperparameters, including the coefficients $\alpha$, $\beta$, and $\gamma$. The model's performance on the test dataset is evaluated using mean absolute error (MAE) as the metric. Specifically, the MAE for each design scheme, generated by the model's design layer, is calculated by comparing the desired mechanical behaviors with the predicted values obtained from the predictive models.

\section{Experimental fabrication, testing, and validation}
\label{sec3}

\subsection{Fabrication }  
\label{sec31}

Regenerated silk fibroin (RSF) was prepared from \textit{Bombyx mori} cocoons following an established method \cite{Rockwood2011}. Briefly, silk cocoons were boiled in a 0.02 M sodium carbonate (Na\textsubscript{2}CO\textsubscript{3}) solution for 30 minutes to remove sericin, then thoroughly rinsed with deionized water to eliminate any remaining chemicals. The degummed silk was dried at room temperature overnight and subsequently dissolved in a 9.3 M lithium bromide (LiBr) solution at 60\textdegree{}C. This solution was dialyzed against deionized water for 72 hours using a 3.5 kDa molecular weight cutoff membrane to remove impurities. The final RSF concentration was determined by calculating the ratio of dry silk fibroin weight to the solution's weight, resulting in a purified RSF solution. For further enrichment, the RSF solution was transferred into the 3.5 kDa molecular weight cutoff membrane to achieve the desired RSF concentration for experimentation.

The RSF solution was carefully transferred into plasma-treated polydimethylsiloxane (PDMS) molds shaped into sinusoidal and dog-bone designs, then trimmed with a glass slide to ensure even surfaces. A dialysis membrane was placed over the structures smoothed to remove any wrinkles. For cylindrical forms, 0.2 mL of silk fibroin solution was dispensed into wells of a 96-well plate, which served as molds. Salt treatments were applied to the dialysis membrane to modify the RSF's structure. Specifically, treatments included 4 M NaCl for 2 hours, 4 M K\textsubscript{2}HPO\textsubscript{4} for 2 hours, and 2.3 M (NH\textsubscript{4})\textsubscript{2}SO\textsubscript{4} overnight. At the salt treatments, the dialysis membrane was carefully removed to release the fabricated silk structures. These were then rinsed in water at 30-minute intervals to eliminate residual salts. 

\subsection{Testing } 
\label{sec32}

\begin{figure*}[ht]
  \centering
  \includegraphics[width=1.0\textwidth]{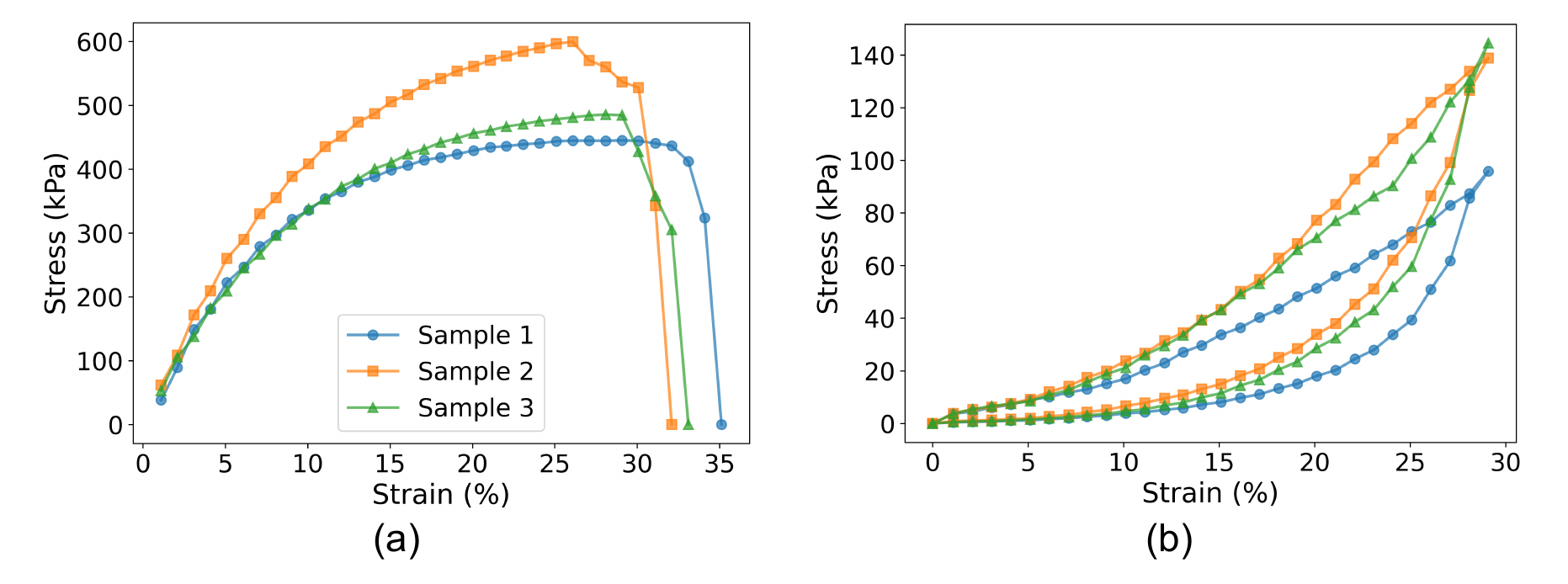}
  \caption{Stress-strain curves from uniaxial (a) tensile and (b) compression tests on 21\% RSF solution samples.}
  \label{fig:testcurves}                
\end{figure*}

Uniaxial tensile tests were performed on a TA Instruments Rheometer H30 device at 25\textdegree{}C with a strain rate of 1\% per second. The dog-bone samples had an estimated thickness of 1 mm and a width of 2 mm. Three samples of 21\% RSF solution were tested, with their stress-strain curves displayed in Figure~\ref{fig:testcurves}(a). Similarly, uniaxial compression tests were conducted using cylindrical samples (6.2 mm diameter) on the same device at 25\textdegree{}C, applying loading and unloading cycles of up to 30\% strain at 1\% per second. Three samples of the same RSF solution were tested, and the corresponding stress-strain curves are shown in Figure~\ref{fig:testcurves}(b). In both figures, two curves closely align, while the third deviates. To ensure a more representative material input for the FEM simulations discussed in Sections~\ref{sec21}, we selected the centrally positioned curve in Figure~\ref{fig:testcurves}(a) and \ref{fig:testcurves}(b). 

\begin{figure*}[ht]
  \centering
  \includegraphics[width=0.75\textwidth]{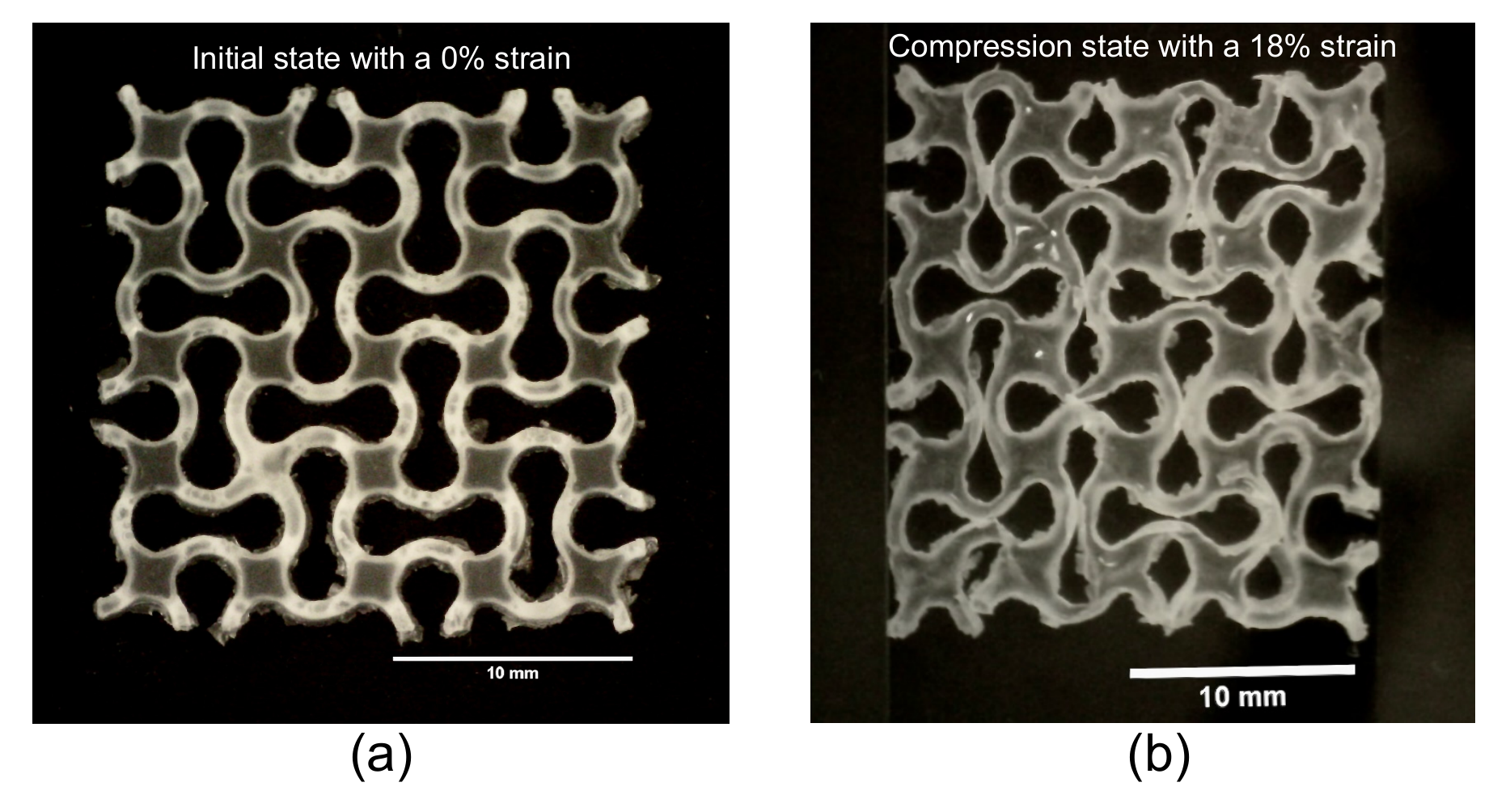}       
  \caption{(a) Undeformed and (b) deformed configurations of a re-entrant sinusoidal auxetic silk fibroin structure.}
  \label{fig:testsamples}                
\end{figure*}

To test the metastructured samples, re-entrant sinusoidal auxetic silk fibroin structures were laterally compressed manually using two glass slides. Compression videos were recorded using a Dino-lite digital microscope and analyzed using ImageJ software. At each strain increment, a box was drawn around the boundary of the structure to measure dimensional changes. These were then used to calculate the Poisson's ratio Eqn~(\ref{eq:poi_define}), relating transverse and axial strain changes. Figure~\ref{fig:testsamples}(a) shows the initial, undeformed configuration, while Figure~\ref{fig:testsamples}(b) shows the deformed configuration under 18\% strain. 

\subsection{Validation}
\label{sec33}

Before proceeding with FEM simulations to generate data for ML, it is crucial to validate the FEM modeling of metastructured patches using experimental outcomes. In simulations conducted in ANSYS, we used the geometry shown in Figure~\ref{fig:testsamples}(a) and the material properties from the tensile and compression tests in Figure~\ref{fig:testcurves}. The simulated compression process closely replicated the experimental setup detailed in Section~\ref{sec32}. Notably, no sliding was observed between the glass slides and the patch during experimental compression. To reproduce this contact behavior in the simulation, we set the friction coefficient to 1.0 to prevent sliding. We used the same element type outlined in Section~\ref{sec21} and calculated Poisson's ratios following the methodology applied in the experiments.

\begin{figure*}[ht]
  \centering
  \includegraphics[width=1.0\textwidth]{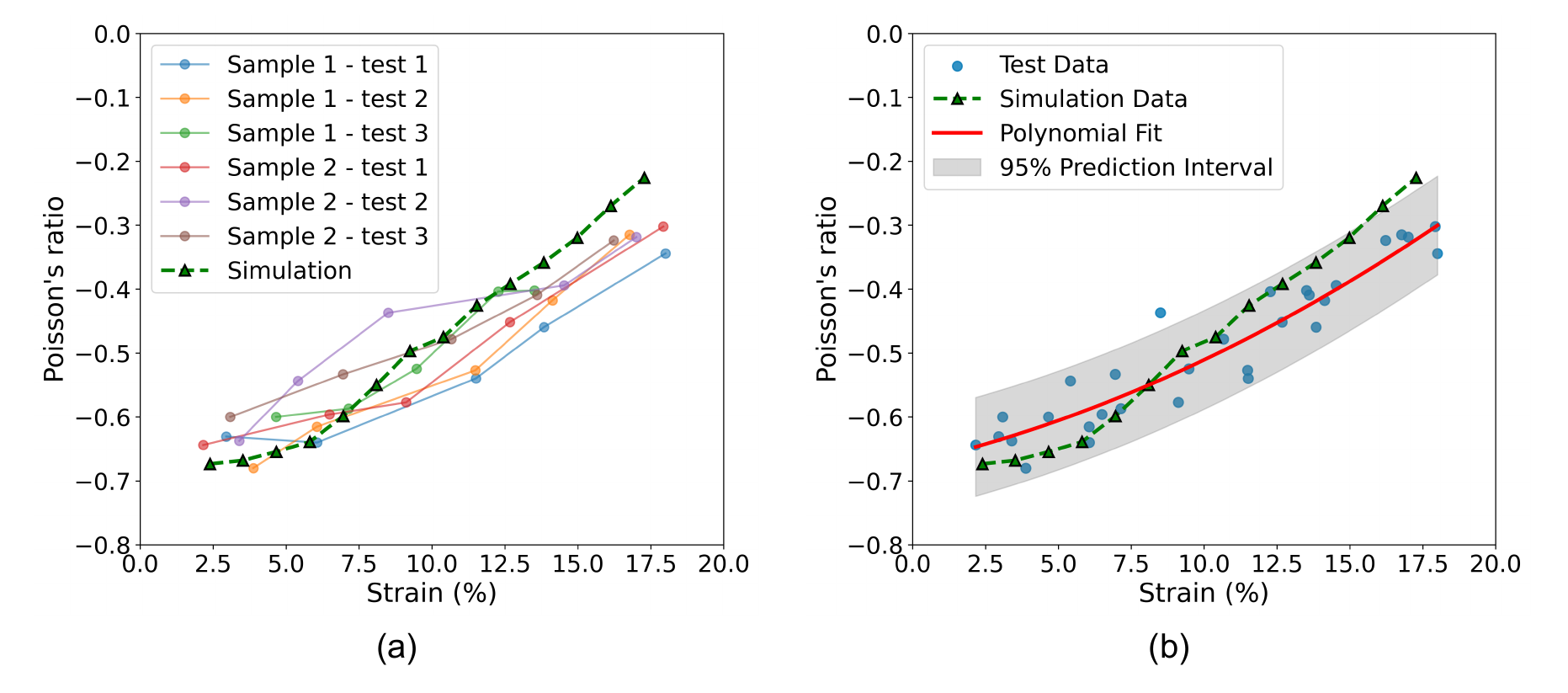}       
  \caption{Comparion of FEM simulation results and experimental measurements of Poisson's ratios at various strain levels. (a) Stress-strain curves from each test and (b) Polynomial regression analysis with a 95\% prediction interval.}
  \label{fig:experimental_val}                
\end{figure*}

As shown in Figure~\ref{fig:experimental_val}, we present both experimental and simulation results. Two samples with identical geometries were fabricated, each tested in compression three times. Figure~\ref{fig:experimental_val}(a) displays stress-strain curves for each test along with the simulation results. Additionally, Figure~\ref{fig:experimental_val}(b) consolidates all test data and includes a second-order polynomial regression with a 95\% prediction interval (PI). It can be observed that the simulation results align within the 95\% PI up to a 15\% strain. However, beyond this strain, the simulation diverges from the PI. This discrepancy likely occurs because the patch undergoes severe squeezing in later compression stages, as shown in Figure~\ref{fig:testsamples}(b). The existence of defects in the actual materials causes some local deformations to be random, making precise calculations challenging. 

\section{Numerical results and discussions}\label{sec4}

\subsection{Mechanical properties prediction}
\label{sec41}

\begin{figure*}[ht]
  \centering
  \includegraphics[width=0.5\textwidth]{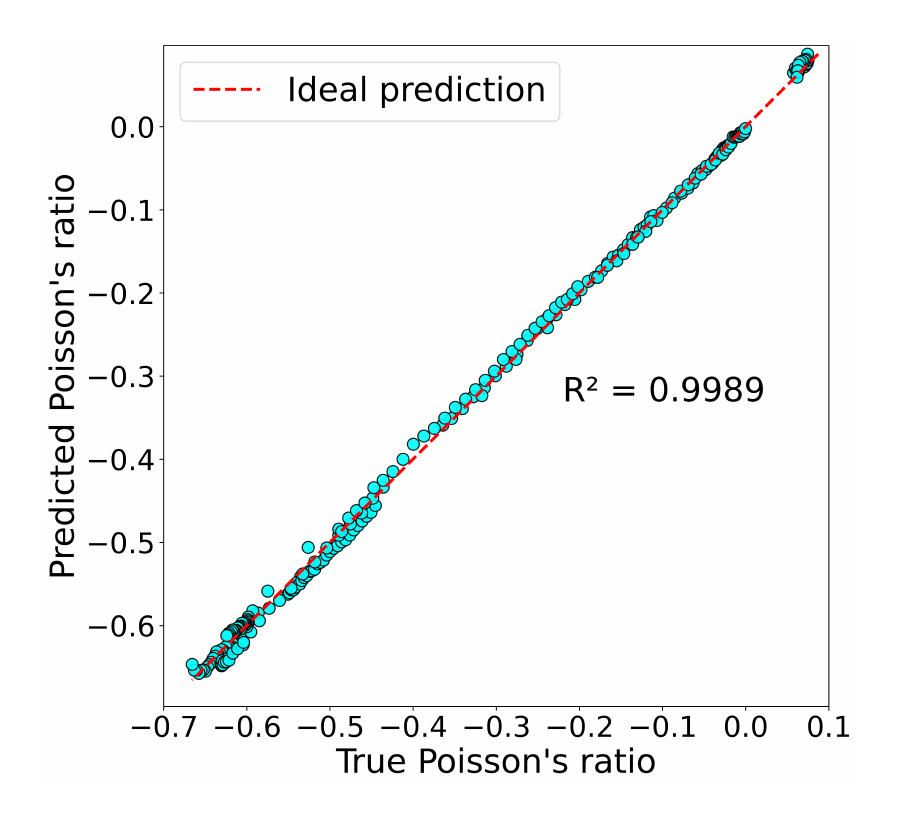}       
  \caption{Comparison of predictions and actual Poisson's ratios.}
  \label{fig:Poi_forward_performance}                
\end{figure*}

We trained two ANN models to predict Poisson's ratios and stresses at various strains, respectively, as described in the previous section. The models were then evaluated on a test set - an unseen dataset - to assess their predictive accuracy. Figure~\ref{fig:Poi_forward_performance} displays the results of the Poisson's ratio predictive model, showing a comparison between predicted and actual Poisson's ratios. This visualization offers a comprehensive view of the model's performance across the entire test dataset. The proximity of data points to the diagonal dashed line, representing ideal prediction, indicates the model's precision. Additionally, an $R^2$ score of 0.9989 highlights the model's high accuracy. 

\begin{figure*}[ht]
  \centering
  \includegraphics[width=1.0\textwidth]{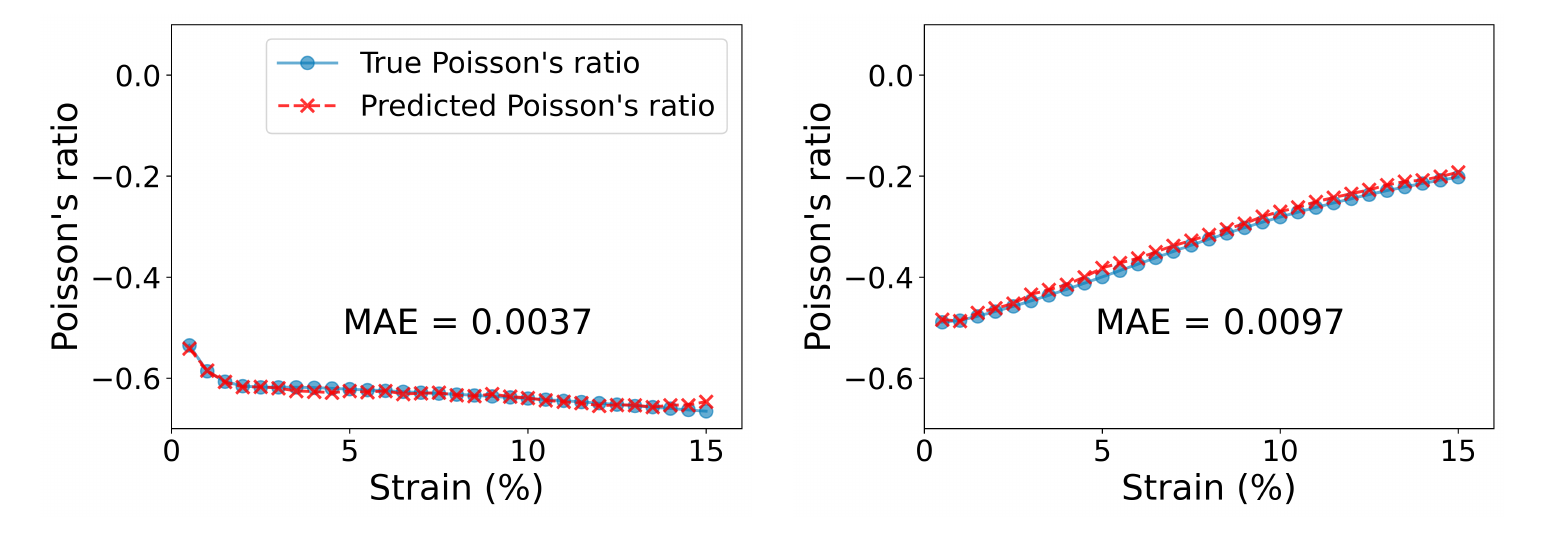}       
  \caption{Predicted Poisson's ratios across various strains for two patch samples in the test set.}
  \label{fig:Poi_forward_samples}                
\end{figure*}

Figure~\ref{fig:Poi_forward_samples} offers an intuitive visualization of the model's predictive capabilities by comparing true and predicted Poisson's ratios across two distinct test data samples. This figure illustrates how closely the model predictions align with the true values over a range of strain levels. The slight divergence between predicted and actual values is quantified by the MAE, which provides a straightforward measure of the average prediction error. A low MAE indicates that the model can reliably estimate Poisson's ratios with minimal deviation, reinforcing its practical utility in applications requiring precise material behavior predictions. 

\begin{figure*}[h]
  \centering
  \includegraphics[width=0.5\textwidth]{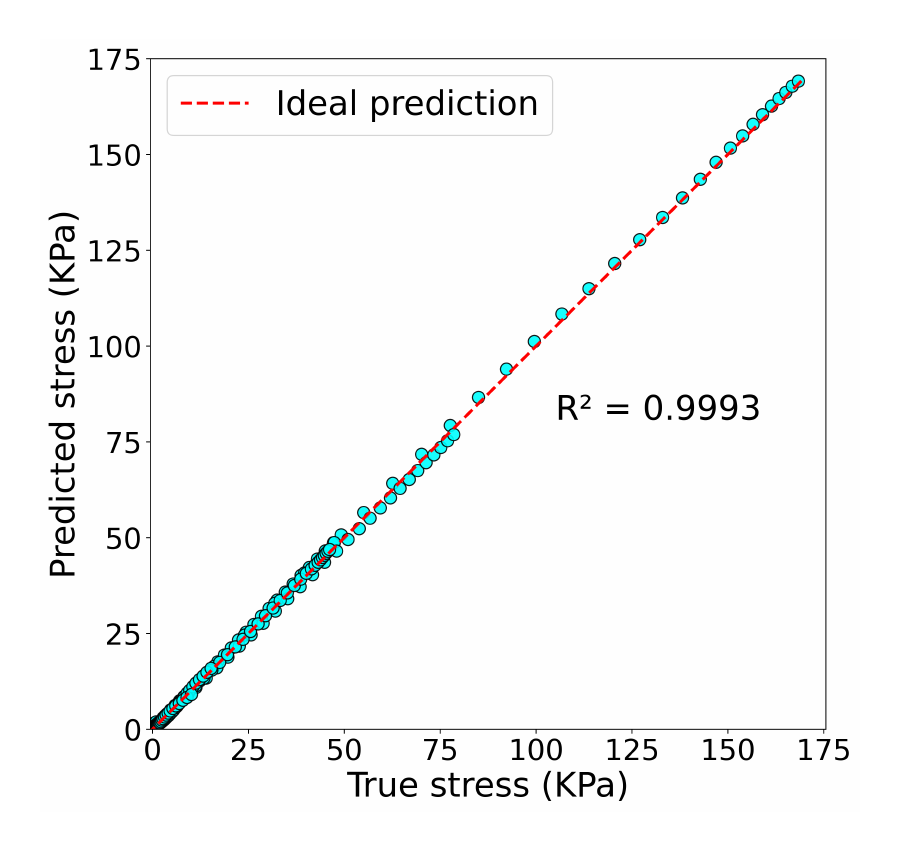}       
  \caption{Comparison of predictions and actual stresses.}
  \label{fig:Stress_forward_performance}                
\end{figure*}

\begin{figure*}
  \centering
  \includegraphics[width=1.0\textwidth]{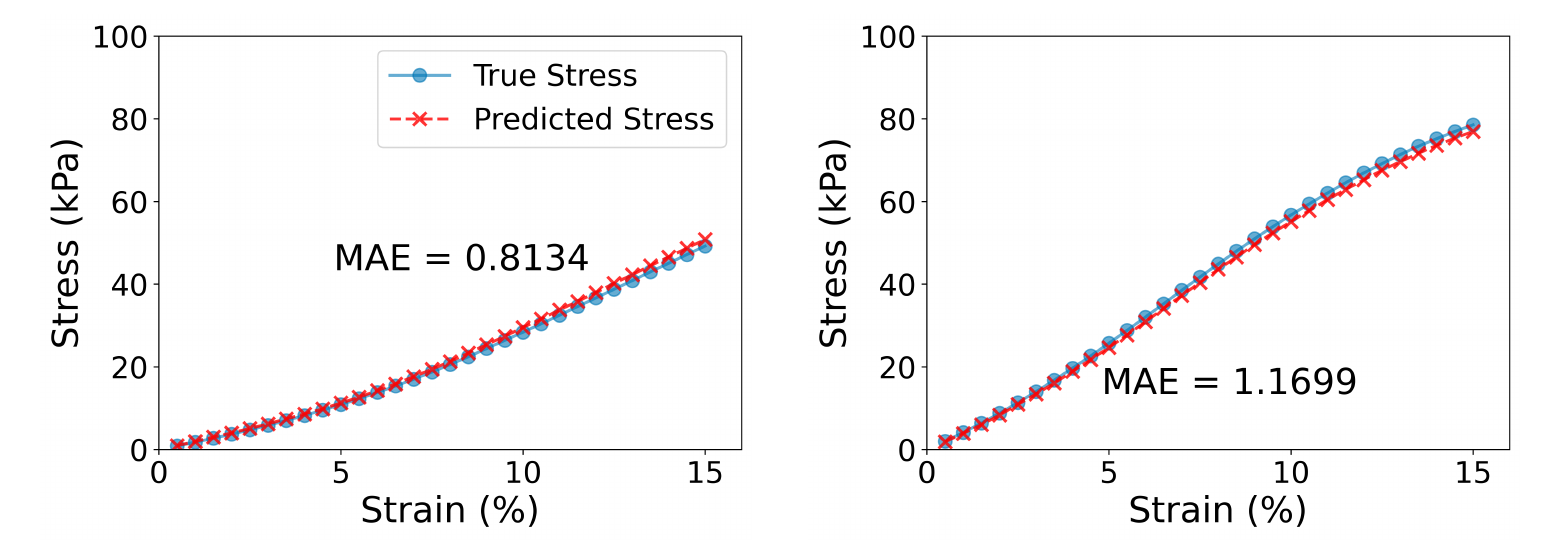}       
  \caption{Predicted stresses across various strains for two patch samples in the test set.}
  \label{fig:Stress_forward_samples}                
\end{figure*}

In addition, the performance of the stress predictive model, as shown in Figure~\ref{fig:Stress_forward_performance} and Figure~\ref{fig:Stress_forward_samples}, further highlights the model's robustness. With an $R^2$ score of 0.9993 on the test dataset, this model demonstrates an exceptional ability to capture complex relationships between stress and strain. The $R^2$ score close to 1 indicates that the model can explain nearly all the variance in the data, which is especially valuable for applications in materials science and engineering where accurate stress predictions are critical. 

The high accuracy of both models is significant for several reasons. Firstly, the models' reliability allows for precise predictions in material design, particularly for simulations that require accurate estimations of mechanical properties across different strain conditions. By reducing dependency on experimental data, these models could accelerate design cycles and improve material selection processes in fields such as aerospace, automotive, and civil engineering. Secondly, the models' predictive success on unseen test samples suggests that they generalize well to new data, making them suitable for integration into digital twins of materials, where real-time predictions of mechanical behavior are essential. Moreover, the low MAE and high $R^2$ scores underscore the ANN models' effectiveness in capturing nonlinear relationships within the data, a task that conventional linear models may struggle to achieve. 

\subsection{Explainability and sensitivity}
\label{sec42}

After training the two predictive models, we used SHapley Additive exPlanations (SHAP) values to analyze the contributions of individual design variables to the predicted mechanical properties. Specifically, we employed the GradientExplainer from the SHAP library in Python, which is well-suited for deep learning models. To provide a reference, we used fifty samples from the training dataset as background data for the explainer. SHAP values were then calculated for the test dataset used in the evaluation of the predictive models.

In addition, we conducted a sensitivity analysis for the design variables $\lambda$, $t$, and $A$. In this analysis, we generated 300 samples. In each set of 100 samples, we kept two variables constant and varied the third at equal intervals within the range used in the GS method. We then used two predictive models to obtain Poisson's ratio and stress data over strains for each configuration, resulting in 100 sets of strain-Poisson's ratio and strain-stress curves for each variable. To quantify sensitivity, we first identified the curve with the minimum mean absolute value as the baseline. Next, we computed the Euclidean distance between this baseline and each of the other 99 curves. By plotting these computed Euclidean distances against the values of the variable being varied, we fitted a linear regression model with a linear basis function to establish the relationship between changes in the variable and deviations in the curves. The slope of this regression model served as the sensitivity, representing how sensitive the output is to variations in each design variable.  

\begin{figure*}[h]
  \centering
  \includegraphics[width=1.0\textwidth]{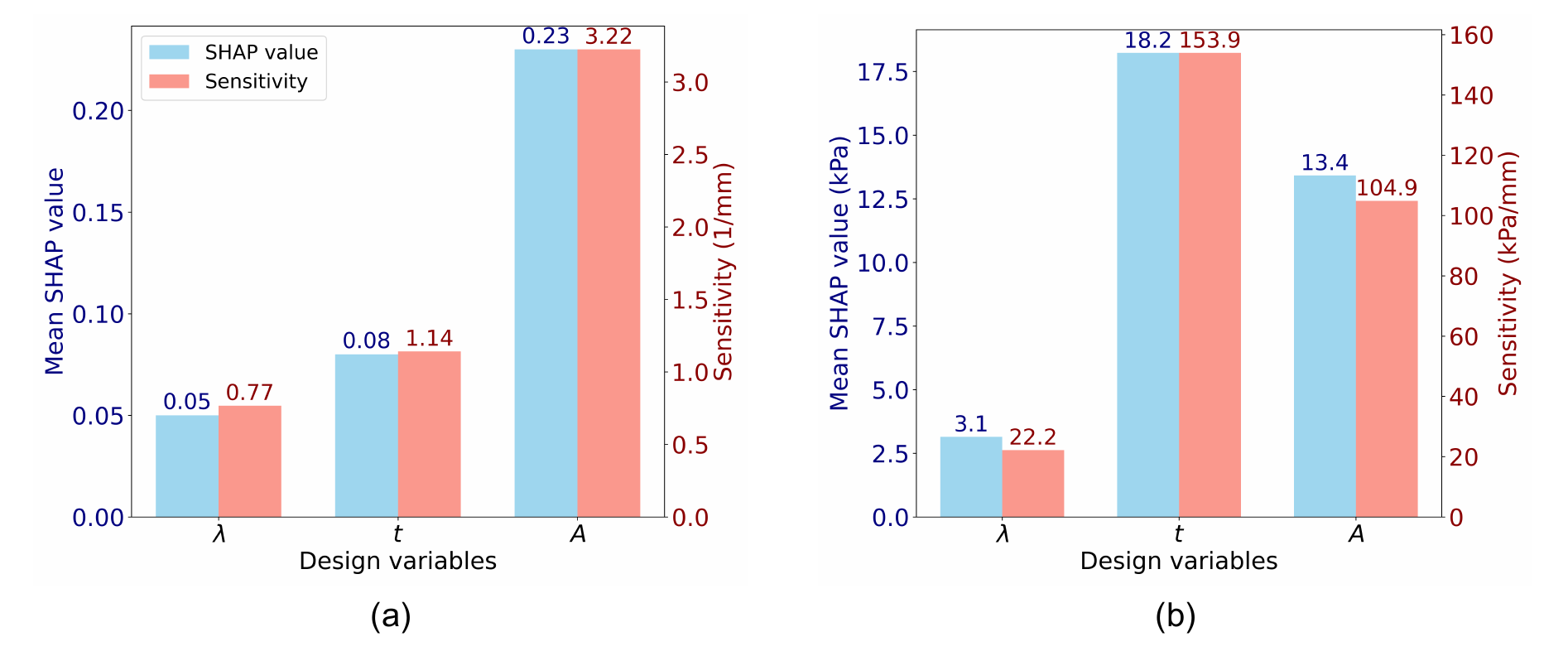}
  \caption{Comparative Analysis of variable influence in (a) Poisson's ratio and (b) stress models using SHAP and sensitivity methods.}
  \label{fig:SHAP_sensitivity}
\end{figure*}

Figure~\ref{fig:SHAP_sensitivity} presents the mean SHAP values and sensitivity analysis results for the three design variables $\lambda$, $t$, and $A$. In this figure, the SHAP values are depicted in light blue, representing the mean absolute SHAP values across all outputs, which provides a comprehensive measure of each variable's overall importance in the model's predictions. A higher SHAP value signifies a greater influence on the model, indicating that the variable plays a critical role in determining the output. Similarly, the sensitivity values, shown in red, highlight how much the model's output is affected by small variations in each design variable. Larger sensitivity values suggest that even minor adjustments to a variable can lead to substantial changes in the output. 

To facilitate meaningful comparison, we normalized the heights of the highest SHAP value and the highest sensitivity in this histogram to be equal. This normalization allows us to assess the relative importance of each variable on the same scale. Figure~\ref{fig:SHAP_sensitivity}(a) displays the analysis results for Poisson's ratio predictive model, revealing a consistent pattern across both analyses: variable $A$ emerges as the most influential, followed by $t$, while $\lambda$ has the least influential. This proportional similarity indicates that the predictive model has effectively captured specific patterns and relationships within the training data. Figure~\ref{fig:SHAP_sensitivity}(b) further corroborates these findings for the stress predictive model, where variable $t$ is identified as the most influential factor affecting stress, with $A$ following closely behind, and $\lambda$ again being the least influential. The distinction between the most influential variables for the two models suggests that different design aspects may govern the behavior of Poisson's ratio and stress under varying conditions.  

The insights gleaned from the SHAP values and sensitivity analysis have practical implications for material design and process optimization. By identifying the key variables that most significantly impact the predictive models, we can prioritize them in future experimental studies or simulations. Moreover, these findings can inform the development of more refined predictive models. By focusing on the most sensitive variables, we can potentially improve model accuracy and robustness, ensuring that the predictive models remain reliable even in the presence of uncertainty or variability in the input parameters. This iterative process of model refinement and validation is essential for advancing our capabilities in predictive analytics and decision-making in engineering contexts.

\subsection{Patch design}
\label{sec43}

To demonstrate the developed design model shown in Figure~\ref{fig:Design_model} for generating patch configurations based on specific mechanical properties, we evaluated it using a randomly sampled test dataset that was not part of the training process. We first assessed the model's capability to produce individual design configurations optimized for given properties. In this scenario, the design layer in the model contains 3 neurons ($n=1$) to yield a single set of design variables. Before being fed into the model, Poisson's ratio and stress values from the training dataset were standardized to have a mean of 0 and a standard deviation of 1. Consequently, we set $\alpha = \beta = 1.0$ and $\gamma = 0$ in the loss function (Eqn~(\ref{eq:loss_define})), reflecting equal priority on accurate predictions for both Poisson's ratio and stress. 

For comparative analysis, we generated design configurations using GA optimization \cite{Mirjalili2019}, a method commonly used to solve optimization and search problems through biologically inspired operators such as selection, crossover, and mutation. In our GA implementation, the three design variables were encoded to a series of binary digits, forming chromosomes. The process started by initiating a diverse population of 100 individuals within the bounds specified in the GS method (Section~\ref{sec21}). The fitness of each individual was evaluated by calculating the MSE between the mechanical properties predicted by the predictive models and the target values. Tournament selection with a size of 2 was employed, with crossover and mutation applied at a rate of 0.8 to produce new candidate solutions. Additionally, elitism was used to retain the best solution from each generation. After 100 iterations, the GA identified the optimal design configuration with the highest fitness score.

\begin{figure*}[ht]
  \centering
  \includegraphics[width=1.0\textwidth]{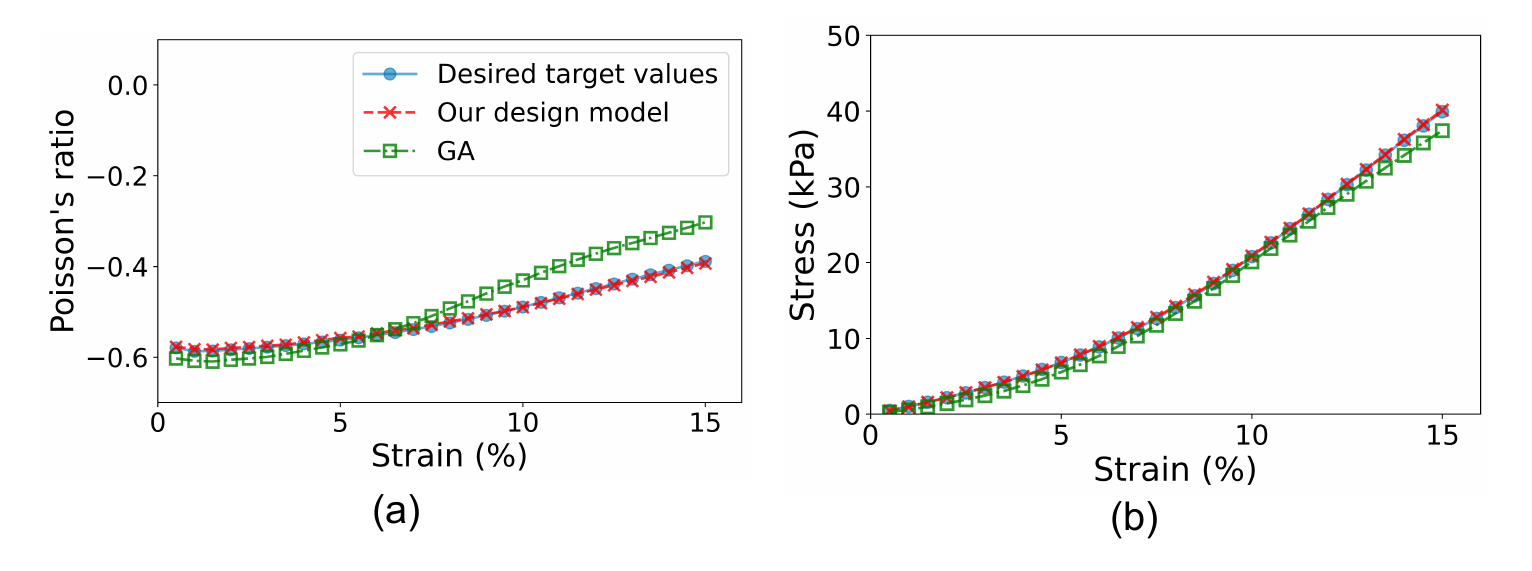}       
  \caption{Comparison of (a) Poisson's ratios and (b) stresses for metastructured patches designed by our design model and GA with desired target values.}
  \label{fig:singe_inverse_2_samples}                
\end{figure*}

Figure~\ref{fig:singe_inverse_2_samples} illustrates the results for a sample from the test dataset. Given specific target mechanical properties, both the design model and GA were used to generate optimal design configurations. These configurations were subsequently employed to predict mechanical properties using the predictive models. The figure shows three curves: the desired mechanical properties and the properties predicted based on configurations generated by the design model and GA, respectively. Figure~\ref{fig:singe_inverse_2_samples}(a) corresponds to Poisson's ratio, while Figure~\ref{fig:singe_inverse_2_samples}(b) represents stress. The results indicate that the configuration produced by GA deviates more from the desired mechanical properties compared to the configuration generated by the design model. 

\begin{table}[h]
\caption{Optimal designs from our design model and GA with corresponding performance metrics.}
\label{table:inverse_model_3_samples}%
\centering
\begin{tabular}{@{}p{0.12in}p{0.15in}p{0.15in}p{0.12in}p{0.15in}p{0.15in}p{0.35in}p{0.35in}p{0.12in}p{0.15in}p{0.15in}p{0.35in}p{0.35in}@{}}
\toprule
\multicolumn{3}{c}{\parbox{0.8in}{True design \\ (mm)}} & \multicolumn{5}{c}{\parbox{1.9in}{Optimal designs from our design model }} & \multicolumn{5}{c}{\parbox{1.9in}{Optimal designs from GA }}\\ 
\cmidrule(lr){1-3} \cmidrule(lr){4-8} \cmidrule(lr){9-13}
$\lambda$ & $t$ & $A$ & $\lambda$ & $t$ & $A$ & \parbox{0.4in}{$\text{MAE}_{\nu}$} & \parbox{0.4in}{$\text{MAE}_{\sigma}$\\(kPa)} & $\lambda$ & $t$ & $A$ & \parbox{0.4in}{$\text{MAE}_{\nu}$} & \parbox{0.4in}{$\text{MAE}_{\sigma}$\\(kPa)} \\ 
\midrule
9 & 1.10 & 0.50 & 9 & 1.12 & 0.51 & 0.0031 & 0.9449 & 9 & 1.14 & 0.50 & 0.0127 & 2.5565 \\
18 & 1.40 & 1.70 & 18 & 1.46 & 1.73 & 0.0032 & 0.0827 & 18 & 1.07 & 1.44 & 0.0416 & 1.0905 \\
14 & 0.90 & 1.00 & 14 & 0.93 & 1.02 & 0.0015 & 0.4130 & 14 & 0.88 & 1.04 & 0.0250 & 1.0056 \\
\bottomrule
\end{tabular}
\end{table}

To present the results more intuitively and quantitatively, Table~\ref{table:inverse_model_3_samples} includes three distinct samples from the test dataset. For each sample, we list both the actual design parameters and those generated by the design model and the GA, respectively. To facilitate direct comparison, we rescaled the designs from the design model and GA so that the variable $\lambda$ matched that of the actual designs. This approach was valid because the mechanical properties remain consistent under proportional scaling. For instance, for the first sample in Table~\ref{table:inverse_model_3_samples}, the design model initially produced $\lambda=23.41\text{mm}$, $t=2.93\text{mm}$, and $A=1.33\text{mm}$. After scaling, these values were adjusted to $\lambda=9\text{mm}$, $t=1.12\text{mm}$, and $A=0.51\text{mm}$, aligning $\lambda$ with the actual design. We then used the MAE to quantify discrepancies between the predicted mechanical properties and the desired ones. In the table, $\text{MAE}_{\nu}$ and $\text{MAE}_{\sigma}$ represent the MAE for Poisson's ratio and stress, respectively. 

It is evident that the design variables generated by our model closely align with the actual configurations, performing better than those produced by GA. Notably, for the second sample, the variables derived from GA show considerable deviation from the true design. This observation is further supported by the comparison of MAE values, which indicate that MAEs from GA are substantially higher than those from the design model. Consequently, the developed design model demonstrates superior performance compared to GA. It is important to note that GA often encounters challenges with suboptimal solutions and lacks diversity in convergence \cite{Xiao2021}. To further assess GA's potential, we increased the population size from 100 to 1000. With this large population, the GA achieved results comparable to those of our design model. However, this improvement was accompanied by a significant increase in computational time. In contrast, once trained, our design model can rapidly generate design configurations with high accuracy, making it more suitable for practical applications.

\begin{table}[h]
\caption{Multiple designs from our design model and GA and their performance.}
\label{table:inverse_model_1_sample}%
\centering
\begin{tabular}{@{}p{0.12in}p{0.15in}p{0.15in}p{0.12in}p{0.15in}p{0.15in}p{0.35in}p{0.35in}p{0.12in}p{0.15in}p{0.15in}p{0.35in}p{0.35in}@{}}
\toprule
\multicolumn{3}{c}{\parbox{0.8in}{True design \\(mm)}} & \multicolumn{5}{c}{\parbox{1.9in}{Multiple designs from our design model}} & \multicolumn{5}{c}{\parbox{1.9in}{Multiple designs from GA}}\\ 
\cmidrule(lr){1-3} \cmidrule(lr){4-8} \cmidrule(lr){9-13}
$\lambda$ & $t$ & $A$ & $\lambda$ & $t$ & $A$ & \parbox{0.4in}{$\text{MAE}_{\nu}$} & \parbox{0.4in}{$\text{MAE}_{\sigma}$\\(kPa)} & $\lambda$ & $t$ & $A$ & \parbox{0.4in}{$\text{MAE}_{\nu}$} & \parbox{0.4in}{$\text{MAE}_{\sigma}$\\(kPa)} \\ 
\midrule
\multirow{3}{*}{18} & \multirow{3}{*}{1.40} & \multirow{3}{*}{1.70} & 18 & 1.49 & 1.74 & 0.0050 & 0.7487 & 18 & 1.07 & 1.44 & 0.0416 & 1.0905 \\
 & & & 18 & 1.26 & 1.55 & 0.0219 & 0.5980 & 18 & 0.96 & 1.42 & 0.0427 & 2.4765 \\
 & & & 18 & 1.16 & 1.50 & 0.0272 & 0.3969 & 18 & 0.90 & 1.31 & 0.0677 & 1.3674 \\
\bottomrule
\end{tabular}
\end{table}

To further explore the diversity of our proposed design model, we set $n=3$ in the design layer to generate three top design configurations that meet the desired mechanical properties. This approach supports real-world applications by offering engineers multiple design options, enabling them to choose the best configuration based on additional considerations. In the loss function, we used coefficients of $\alpha=1.0$, $\beta=1.0$, and $\gamma=0.5$. 

Table~\ref{table:inverse_model_1_sample} presents a comparative analysis using the second sample from Table~\ref{table:inverse_model_3_samples}. For GA optimization, we sorted the fitness values of the final population and selected the top three individuals as the distinct design configurations. Table~\ref{table:inverse_model_1_sample} shows that the MAE values for designs generated by our model remain impressively low, even with multiple candidates. However, when comparing the optimal design (first design ) from the multi-design model in Table~\ref{table:inverse_model_1_sample} with the corresponding single-design model configuration (second sample) in Table~\ref{table:inverse_model_3_samples}, a slight reduction in accuracy is noted. This can be attributed to the inclusion of specific loss term $L_{\text{scale}}(\theta)$, which enhances diversity at the cost of a small decrease in precision. Despite this minor trade-off, Table~\ref{table:inverse_model_1_sample} effectively demonstrates the design model's capability to produce diverse, highly accurate solutions.

\subsection{Limitations}
\label{sec44}

In the studies above, both the training and test datasets for the design model were generated through FEM simulations of sinusoidal metastructures with varying design configurations. We hypothesized that the model's generalization capability might be constrained due to the limited design diversity in the training data. To explore this potential limitation, we tested the single-design model previously developed, using the same training set but introducing a new objective: designing a sinusoidal metastructure that could replicate the mechanical properties of an oval voids structure. Specifically, the target mechanical properties - Poisson's ratios and stresses - were derived through FEM modeling and simulation of a randomly generated oval voids metastructure. This approach enabled us to assess whether the design model could adapt and generate a sinusoidal metastructure capable of reproducing the distinct mechanical properties observed in an oval voids structure.

\begin{figure*}[ht]
  \centering
  \includegraphics[width=0.95\textwidth]{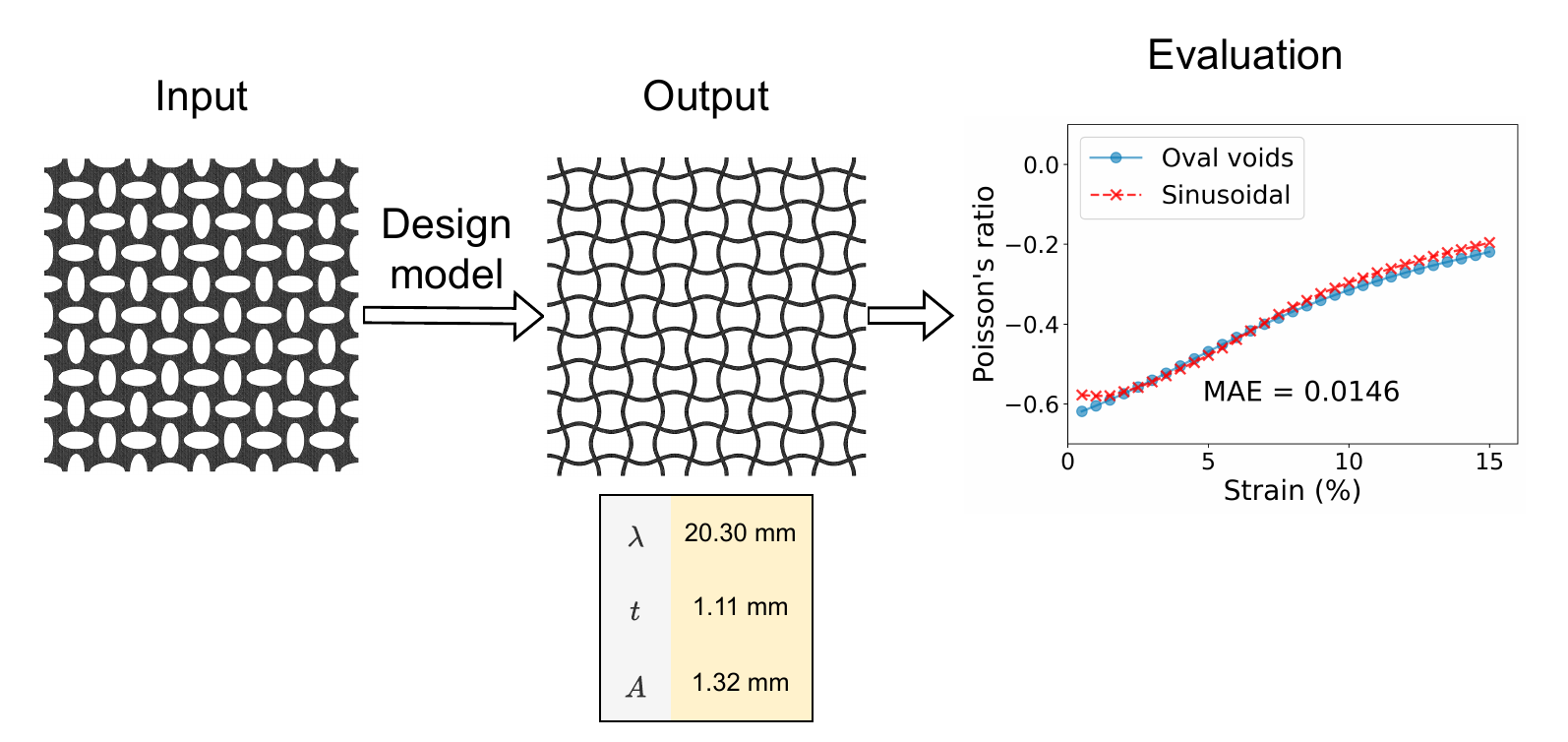}       
  \caption{Design of a sinusoidal metastructure to replicate Poisson's ratios of an oval voids structure across various strain levels.}
  \label{fig:inverse_generalization_poi}                
\end{figure*}

\begin{figure*}[ht]
  \centering
  \includegraphics[width=0.95\textwidth]{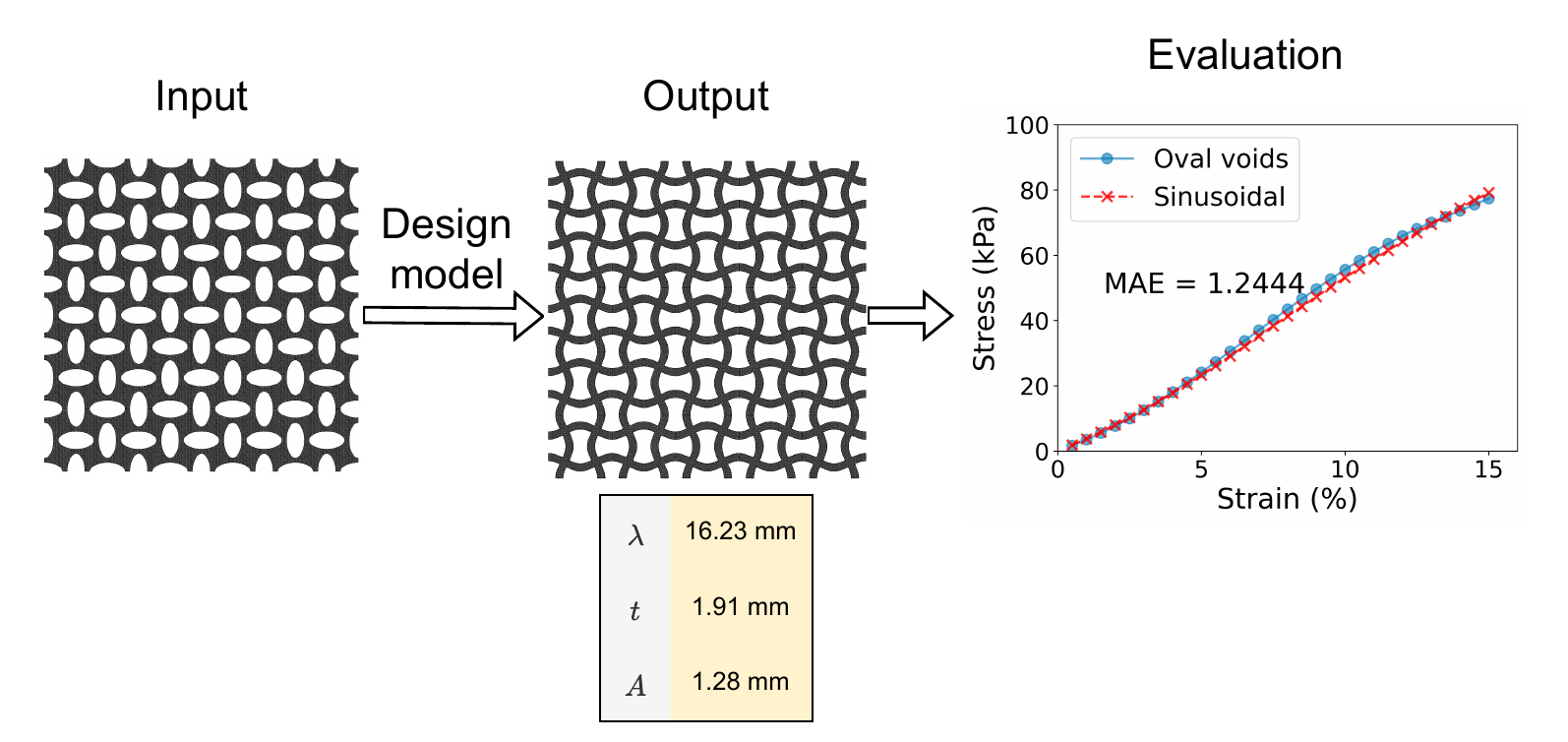}       
  \caption{Design of a sinusoidal metastructure to replicate stresses of an oval voids structure across various strain levels.}
  \label{fig:inverse_generalization_stress}                
\end{figure*}

\begin{figure*}[h]
  \centering
  \includegraphics[width=0.95\textwidth]{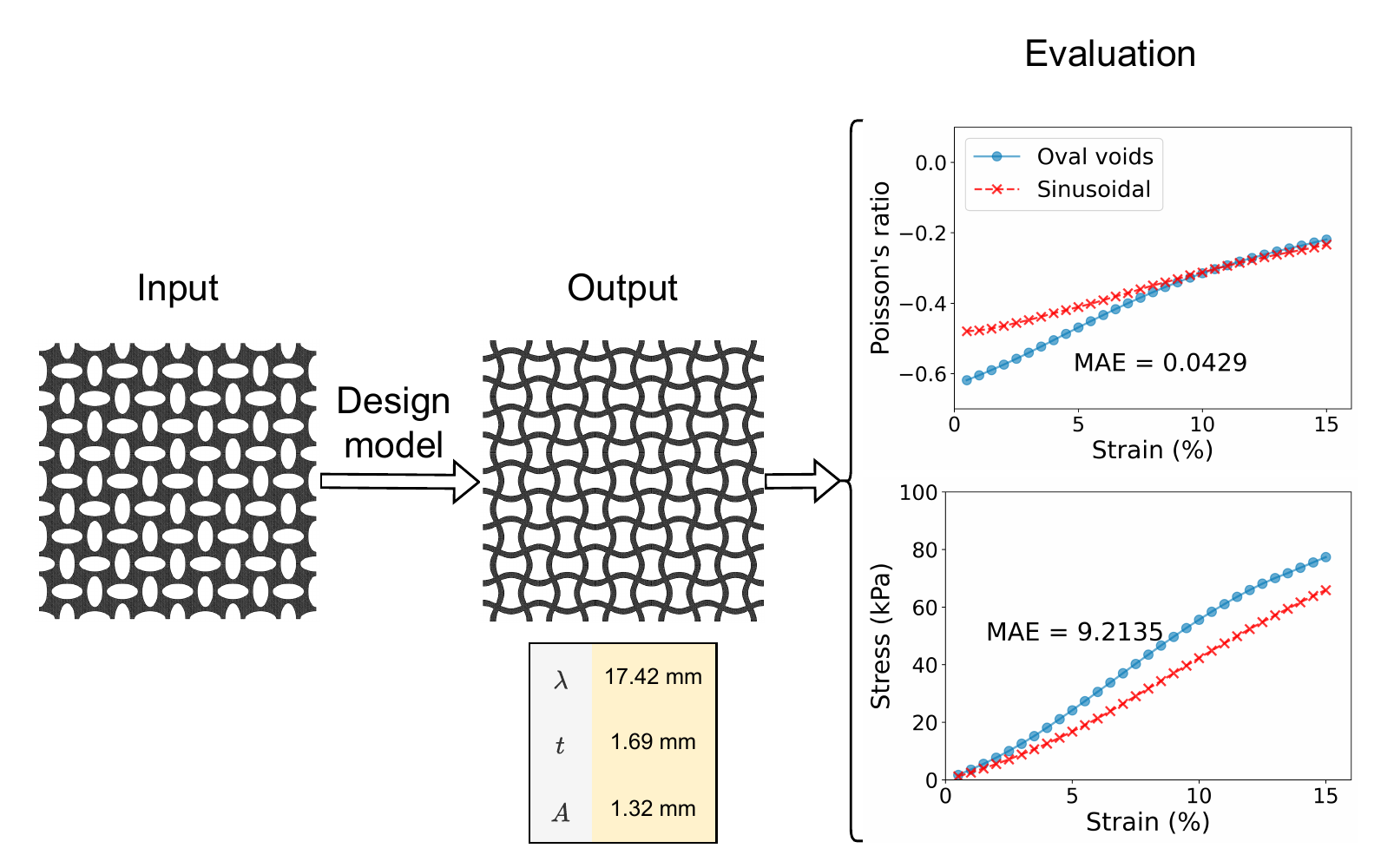}       
  \caption{Design of a sinusoidal metastructure to replicate both mechanical properties of an oval voids structure across various strain levels.}
  \label{fig:inverse_generalization_both}                
\end{figure*}

We conducted initial tests on Poisson's ratio and stress separately. For Poisson's ratio, the design model was trained using a loss function with the weights of $\alpha=1.0$ and $\beta=0$, prioritizing accuracy for Poisson's ratio alone. The design process, as illustrated in Figure~\ref{fig:inverse_generalization_poi}, successfully produced a sinusoidal metastructure that closely matched the target Poisson's ratios, achieving an MAE of 0.0146. For stress, we adjusted the loss function by setting $\alpha=0.0$ and $\beta=1.0$ to focus exclusively on reproducing the stress values. Figure~\ref{fig:inverse_generalization_stress} depicts the resulting design process, wherein the desired stress values were derived from the same oval voids structure as in the Poisson's ratio experiment. Although the sinusoidal metastructure generated in this case differed from that in Figure~\ref{fig:inverse_generalization_poi}, it effectively replicated the stress values with the MAE of 1.2444. 

However, when both Poisson's ratio and stress were simultaneously targeted, as depicted in Figure~\ref{fig:inverse_generalization_both}, the design model was unable to generate a single sinusoidal metastructure that accurately reproduced both mechanical properties. Unlike in the previous cases where each property (Poisson's ratio or stress) was addressed individually, the model struggled to achieve alignment with the target values for both properties, resulting in noticeably higher MAE values. This discrepancy suggests that the sinusoidal design, characterized by only three parameters - $\lambda$, $t$, and $A$ - lacks the geometric versatility required to capture the complex and multifaceted mechanical behavior of different types of metastructures, such as oval voids structures. The simplified sinusoidal geometry does not offer the DOF necessary to replicate both Poisson's ratio and stress simultaneously, as the oval voids structure does with its more intricate geometry and flexibility. An alternative solution could involve adding more training data that include a variety of metastructure types and subsequently updating the architecture of the design model accordingly. 

\section{Conclusions and future works}\label{sec5}

In this study, we developed a novel data-driven framework to predict the mechanical properties of and design bio-inspired auxetic patches fabricated from silk fibroin. By integrating FEM simulations, experimental validation, and neural networks, we effectively addressed both the prediction and design challenges. Our research focused on re-entrant sinusoidal metastructures characterized by three design variables, facilitating efficient sample generation and data collection. We employed the GS active learning technique to reduce the training dataset size, enabling the training of high-performance ML models with only 150 samples.

We developed two ANN models to predict Poisson's ratio and stress at various strain levels up to 15\%. Both models demonstrated excellent performance, achieving $R^2$ scores exceeding 0.995 on unseen test data. Using SHAP values and sensitivity analysis, we gained insights into the influence of individual design variables, with amplitude ($A$) and thickness ($t$) emerging as key factors in determining Poisson's ratios and stresses, respectively. Additionally, we introduced an innovative ANN-based design model that generates metastructure designs to meet specified mechanical targets. This model outperformed traditional optimization methods, such as GA, by providing more accurate and efficient design solutions. The ability to generate multiple design configurations offers practical flexibility, enabling users to select optimal designs based on additional criteria.

However, our approach has some limitations. Since the design model was trained exclusively on sinusoidal metastructures, it was unable to generate a sinusoidal design that could simultaneously replicate both Poisson's ratio and stress characteristics of different metastructures, such as those with oval voids. This limitation underscores the geometric constraints inherent in the sinusoidal design, which has only three parameters and may not capture the complexity required for more intricate metastructures.

To address these challenges and advance the field, future research should focus on expanding the training dataset to encompass a broader range of metastructure types with more complex geometries. Incorporating a GAN into the design model will enable more flexible and diverse metastructure designs. This expansion would improve the model's ability to generalize and create patches that replicate a wide range of mechanical behaviors. Additionally, integrating multiscale design variables into the metastructures could further enhance the design model's generalization capability. For instance, replacing the bulk material in the patch structure with another metastructure or dividing the patch into multiple layers with distinct materials would increase design flexibility. These approaches would allow for precise customization of mechanical properties to meet specific application needs.

\backmatter

\bmhead{Acknowledgements} 

This research was funded by the US Department of Education (ED\#P116S210005). YC and SX are also supported by the National Science Foundation (\#2104383). XM and MA are supported by Cystic Fibrosis Foundation (006655I223), 2023 UIOWA Jumpstart Tomorrow Pilot award, and the Carver Charitable Trust.

\bmhead{Author contributions} 

Yingbin Chen: Data curation, Formal analysis, Investigation, Methodology, Software, Validation, Visualization, Writing -- original draft. Milad Arzani: Experimental testing, Data curation, Investigation, Validation, Visualization, Writing -- original draft. Xuan Mu: Conceptualization, Formal analysis, Funding acquisition, Methodology, Project administration, Resources, Supervision, Writing -- review \& editing. Sophia Jin: Data curation, Software, Validation. Shaoping Xiao: Conceptualization, Formal analysis, Funding acquisition, Methodology, Project administration, Resources, Supervision, Writing – original draft, Writing -- review \& editing.

\bmhead{Data availability}

The datasets and code generated during the current study are available from the corresponding author upon reasonable request.

\section*{Declarations}

\bmhead{Conflict of interest}

The authors declare that they have no conflict of interest.


\bibliography{Manuscript.bib}

\end{document}